# Using qualia information to identify lexical semantic classes in an unsupervised clustering task

*Lauren ROMEO[1] Sara MENDES[1,2] Núria BEL[1]*

(1) Universitat Pompeu Fabra
Roc Boronat, 138, Barcelona, Spain

(2) Centro de Linguística da Universidade de Lisboa
Avenida Professor Gama Pinto, 2, Lisboa, Portugal

`{lauren.romeo,sara.mendes,nuria.bel}@upf.edu`

ABSTRACT

Acquiring lexical information is a complex problem, typically approached by relying on a number of contexts to contribute information for classification. One of the first issues to address in this domain is the determination of such contexts. The work presented here proposes the use of automatically obtained FORMAL role descriptors as features used to draw nouns from the same lexical semantic class together in an unsupervised clustering task. We have dealt with three lexical semantic classes (HUMAN, LOCATION and EVENT) in English. The results obtained show that it is possible to discriminate between elements from different lexical semantic classes using only FORMAL role information, hence validating our initial hypothesis. Also, iterating our method accurately accounts for fine-grained distinctions within lexical classes, namely distinctions involving ambiguous expressions. Moreover, a filtering and bootstrapping strategy employed in extracting FORMAL role descriptors proved to minimize effects of sparse data and noise in our task.

# 1 Introduction

Acquiring lexical information is a complex problem, typically approached by relying on a number of contexts to contribute information for classification, following the Distributional Hypothesis (Harris, 1954) and the idea of distributional similarity. In this domain it is crucial to determine which distributional information is significant to characterize lexical items. In line with Pustejovsky and Ježek (2008), we will make apparent how focusing on occurrences indicative of the FORMAL role of the Generative Lexicon (GL) theory (Pustejovsky, 1995) allows for identifying lexical semantic classes.

Lexical classes are linguistic generalizations regarding characteristics of meaning that correspond to sets of properties shared by groups of words. Bybee and Hopper (2001) and Bybee (2010) state that words are organized in lexical-semantic classes defined as emergent properties of words that recurrently occur in a set of particular contexts. Though many NLP tasks rely on rich lexica annotated with lexical semantic classes, reliable lexical resources including this type of lexical information are mostly manually developed, which is unsustainable, costly and time-consuming, and makes conceiving methods to automatically acquire such information crucial. An approach for acquiring lexical semantic classes proposes to classify words according to their occurrences in contexts where other lexical items belonging to a known class also occur. Yet, this approach has some limitations, such as data sparseness and noise (see Section 2), which underline the importance of developing new strategies to improve its effectiveness. Authors such as Pustejovsky and Ježek (2008) have shown how distributional analysis and theoretical modeling interact to account for rich variation in linguistic meaning. In line with this proposal, we evaluate the significance of specific co-occurrences whose selection was motivated by aspects of GL.

This work attempts to evaluate whether information provided by qualia roles, in specific the FORMAL role, is sufficient to discriminate lexical semantic classes of English nouns. With the experiments depicted in this paper, we aim to empirically demonstrate to which extent these features draw together nouns from the same lexical semantic class in an unsupervised clustering task. In this paper, Section 2 depicts background and motivation of this work. Section 3 presents relevant information on the GL and dot-objects. Section 4 describes the methodology to automatically obtain and cluster FORMAL role descriptors of nouns. Section 5 and 6, respectively, describe and discuss results. Section 7 reflects upon lexical classes and logical polysemy and is followed by final remarks.

# 2 Background and Motivation

Mainstream approaches to lexical semantic class acquisition classify words according to occurrences, i.e. they use the entire set of occurrences of a word to determine class membership. Yet, this approach has some limitations. Blind-theory distributional approaches have been shown to fail to account for the wide range of linguistic behavior displayed by words in language data (see Pustejovsky and Ježek (2008)), while authors such as Bel et al. (2010) reported problems caused by sparse data, or lack of evidence, and noise, or information obtained though not aimed at. Concerning sparse data in classification tasks, nouns that appear only once or twice in a corpus, and not in sought contexts, can render ineffective any classifier or clustering algorithm by not providing sufficient information for classification. We aim to soften effects of sparse data in the context of a clustering task by using a bootstrapping technique reliant on natural language inference properties (see Section 4.1). Noise, another pervasive issue in lexical semantic class acquisition, can be due to different factors: the occurrence of very general nominal expressions (e.g. "kind of"), which do not provide distinguishing lexical information; misleading corpus features; and the use of low-level tools (see Bel et al. (2012)). We assume noise resulting from errors generated by NLP tools to be typically characterized by unique occurrences and we employ a filtering strategy to overcome its possible effects (see Section 4.1). Concerning misleading corpus features, these are often caused by ambiguity of lexical items, resulting in nouns occurring in contexts not corresponding to their assumed lexical class. This presents challenging problems in classification tasks, as most authors do not distinguish among related senses of the same word, i.e. they either consider it as part of the class or not (Hindle, 1990; Bullinaria, 2008; Bel et al., 2012). This is particularly problematic when words allow for multiple selection, i.e. when different senses of the same lexical item can be simultaneously selected for in one sentence (see (1)). Known as logical polysemy, this type of ambiguity has been shown to have well-defined properties (see Pustejovsky (1995) and Buitelaar (1998)) and has been consistently reported as a factor in lexical semantic acquisition tasks.

*The newly constructed* (LOCATION) *bank offers special conditions* (ORGANIZATION) *to new clients.* (1)

Approaches in this domain have usually tried to distinguish and isolate each word sense. We address this phenomenon differently, considering polysemous nouns as members of a given ambiguity class (within a wider lexical semantic class) and making apparent the relation between members of different classes by identifying shared properties beyond class limits. Given these considerations, we assume lexical units are complex objects that display rich variations of meaning in language use, placing ourselves within a theoretical framework that provides us the tools to account for this fact. Using the levels of representation and generative mechanisms in GL, we attempt to soften the effects of the aforementioned limitations in the automatic acquisition of lexical information.

## 3 Generative Lexicon theory

GL models the internal structure of lexical items in a computational perspective (Pustejovsky, 1995), proposing various levels of representation to semantically represent words, while allowing for the computation of meaning in context. Qualia Structure (QS) is one of these levels, consisting of 4 roles (FORMAL: what an object is; CONSTITUTIVE: what it is composed of; TELIC: its purpose; AGENTIVE: its origin), which model the predicative potential of lexical items. Here, we focus on the FORMAL role, defined as the role that distinguishes a lexical object within a larger domain (Pustejovsky, 1991).

QS also models phenomena such as polysemy of lexical items inherently complex in their meaning. These instances, *dot objects*, are the logical pairing of senses denoted by individual types in a complex type (Pustejovsky, 1995), which can pick up distinct aspects of the object, as well as properties of more than one class (Pustejovsky and Ježek, 2008), typically allowing for multiple selection (see (1)). Being able to represent lexical items as complex objects is useful in the context of our work as it provides a formal explanation for words belonging to more than one type, and essentially to more than one class.

Our experiment uses FORMAL role information as features for identifying lexical class membership. However, as there are no lexica available annotated with such information, we needed to obtain it automatically. Automatically extracting qualia roles with lexico-syntactic patterns has been receiving considerable attention for its success: Hearst (1992) identified lexico-syntactic patterns to acquire noun hyponyms, corresponding to the FORMAL role, whereas Cimiano and Wenderoth (2007) identified lexico-syntactic patterns to obtain information regarding semantic relations that correspond to each qualia role. As we needed information regarding the FORMAL role, not full lexical entries, in order for clusters to emerge, following Celli and Nissim (2009), we bypassed the representation of the entire QS, assuming semantic relations can be induced by matching lexico-syntactic patterns that convey a relation of interest.

## 4 Methodology

Given the unavailability of lexica annotated with FORMAL role information, and considering our basic goal of evaluating whether this information is enough to cluster together nouns of the same class, we extracted it from a corpus using lexico-syntactic patterns, following Cimiano and Wenderoth (2007), and then used it as features for a clustering task. In the experiment performed, we employed two steps: the extraction of FORMAL role descriptors from corpus data; and the clustering of this information. To obtain FORMAL role descriptors for our unsupervised clustering task, we used a part of the UkWaC Corpus (Baroni et al., 2009), consisting of 150 million tokens. We employed 60 seed nouns pertaining to three lexical semantic classes: HUMAN, LOCATION, and EVENT. The seed nouns were said to belong to a class if they contained a sense in WordNet (Miller et al., 1990) corresponding to one of the three classes. Seed nouns were not contrasted with actual occurrences in the corpus.

## 4.1 Extraction of FORMAL role descriptors using lexico-syntactic patterns

Firstly, seed nouns were used in handcrafted lexico-syntactic patterns, adapted from Hearst (1992) patterns and the list proposed by Cimiano and Wenderoth (2007), to extract FORMAL role descriptors. These patterns were specified through regular expressions with PoS tags given after each token.

| |
|---|
| $x$_(or/and)_other_$y$ |
| $x$_such_as_$y$ |
| $x$_(is/are)_(a/an/the)_(kind(s)/type(s))_of_$y$ |
| $x$_(is/are)_also_known_as_$y$ |

TABLE 1 – Clues on which patterns used to detect FORMAL role information in corpus data were built

The information obtained was stored in vectors representing co-occurrences with seed nouns in relevant contexts (patterns), where each element corresponds to occurrences of a particular seed noun ($x$) with a possible FORMAL role descriptor ($y$), following Katrenko and Adriaans (2008). Using the clues in Table 1, we obtained 185 FORMAL role descriptors for 55 of the 60 seed nouns in 353 occurrences. Considering this, and given the properties of the clustering algorithm used (see Section 4.2) a random value would be provided to nouns not sharing feature information with any other noun in our data set. To avoid random cluster assignments and provide more significant information to the system, we filtered out the features not shared between at least two seed nouns, without controlling which class the shared features belonged to, thus maintaining an unsupervised environment. Though we employed a large set of data, there were not enough shared FORMAL role descriptors for an important part of our data set, leading us to devise a strategy to increase the information available to the clustering algorithm.

a. A mammal is a [type of] animal.
b. A zebra is a [type of] mammal.
c. Therefore, a zebra is a [type of] animal. (2)

To increase the amount of FORMAL role descriptors, we employed a bootstrapping technique (Hearst, 1998) relying on monotonic patterns for natural language inference (Hoeksema, 1986; van Behthem, 1991; Valencia, 1991), illustrated in (2). This strategy is consistent with GL lexical inheritance structure (Pustejovsky, 1995; 2001), which assumes lexical items obtain their semantic representation by accessing a hierarchy of types and inheriting information according to their QS, meaning qualia elements are viewed as categories hierarchically organized. To illustrate how this applies in our case, the HUMAN noun *treasurer* obtained *officer* as a FORMAL role descriptor, whereas *officer* extracted *person* and *employee* as its own FORMAL role descriptors. Assuming this lexical organization, we consider FORMAL role descriptors extracted for *officer* to also be features of *treasurer*. Thus, we gathered additional information regarding the nouns to cluster, using originally obtained FORMAL role descriptors as "seed nouns" to extract more elements in an attempt to overcome biases due to sparse data (see Section 6), as well as to reinforce information already obtained. Employing the original patterns and original extractions as seeds, we obtained information that was added to the vectors. We conducted one iteration of the bootstrapping technique, going up one level of generalization to obtain the final distribution of information below. Newly obtained information was unified with previously extracted features, filtering out any additional noise attained. Table 2 presents the final distribution of this information.

| Class | Elements | Occurrences |
|---|---|---|
| HUMAN | 61 elements | 841 occurrences |
| LOCATION | 43 elements | 225 occurrences |
| EVENT | 36 elements | 216 occurrences |

TABLE 2 – Distribution of FORMAL role descriptors extracted (after filtering and bootstrapping) per class of seed noun

### 4.1.1 Error Analysis

Basing our clustering experiment on automatically extracted FORMAL role descriptors, the accuracy of the information obtained was a concern. To assess the accuracy of the information obtained, the FORMAL role descriptors extracted were revised manually. Extractions were considered erroneous if they provided information not in accordance with the class that the seed nouns pertained to. Table 3 presents the results of this analysis. Erroneous extractions were due to faults of the extraction mechanism (i.e. problems handling phenomena such as PP attachment), PoS tagging errors, lexical ambiguity or erroneous statements in text (Katrenko and Adriaans, 2008), as well as errors due to logical polysemy (see Section 6). Note that although errors were identified, they were not filtered for the clustering task, i.e. all information (erroneous or not) was included (on the impact of errors in results see Section 6).

| Class | % of accurate FORMAL role descriptors extracted |
|---|---|
| HUMAN | 87.60% |
| LOCATION | 63.54% |
| EVENT | 75.96% |

TABLE 3 – Percentage (%) of accurate FORMAL role descriptors obtained per class

## 4.2 Clustering nouns using FORMAL role information

The second step of our experiment consisted in clustering nouns using the FORMAL role descriptors extracted. Given the nature of our data, we selected the sIB clustering algorithm (see Slonim et al. (2002) for a formal definition) for the manner it manages large data sets. This algorithm calculates similarity between two vectors using the *Jensen-Shannon* divergence, which measures similarity between probability distributions, rather than the Euclidean distance, which can bias the results when the number of attributes representing the factors is unequal (Davidson, 2002). This was our case as our feature spaces depend on the number of FORMAL role descriptors each seed noun occurred with in the corpus. To empirically demonstrate to which extent FORMAL role descriptors draw together nouns from the same class, we designed an experiment using the sIB algorithm in WEKA (Witten and Frank, 2005) to cluster seed nouns into lexical semantic classes, based only on the FORMAL role information obtained.

## 5   Results

As mentioned, our goal was to cluster together nouns from the same lexical semantic class using only FORMAL role descriptors. As the evaluation of unsupervised distributional clustering algorithms is usually done by comparing results to manually constructed resources (see Rumshsiky et al. (2007), among others), we employed our list of pre-classified seed-words to determine if nouns of the same class clustered together. Tables 4 and 5 present clustering results. The distribution of nouns across each cluster is given by the percentage of nouns pertaining to each lexical class included in it. The total number of seed nouns in each cluster is also given.

| Cluster 0 | Cluster 1 | Cluster 2 | Class |
|---|---|---|---|
| **0.9285** | 0 | **0.5714** | HUMAN |
| 0.0769 | 0.3913 | 0.1429 | LOCATION |
| 0 | **0.6087** | 0.2857 | EVENT |
| 14 | **23** | 7 | TOTAL NUMBER OF SEED NOUNS PER CLUSTER |

TABLE 4 – Distribution of nouns in a 3-way clustering solution

We experimented with a 3-way and a 4-way clustering solution. In the first, the number of clusters was defined by the number of known classes, and resulted in the clustering of HUMAN nouns (Cluster 0). LOCATION and EVENT nouns grouped together in Cluster 1, the remaining cluster being composed of nouns from all classes with very few features available (less than three), i.e. insufficient information for classification. Considering this, we employed a 4-way solution to see whether LOCATION and EVENT nouns could be discriminated. This solution distinguished between the three classes (Cluster 0, 1 and 3 in Table 5) with a fourth cluster containing the "sparse data" nouns also affecting the 3-way solution.

| Cluster 0 | Cluster 1 | Cluster 2 | Cluster 3 | Class |
|---|---|---|---|---|
| 0 | 0 | **0.5714** | **0.9286** | HUMAN |
| 0 | **0.9** | 0.1429 | 0.0769 | LOCATION |
| **1** | 0.1 | 0.2857 | 0 | EVENT |
| 13 | 10 | 7 | 14 | TOTAL NUMBER OF SEED NOUNS PER CLUSTER |

TABLE 5 – Distribution of nouns in a 4-way clustering solution

The results show that even after filtering and bootstrapping the features extracted, sparse data still affected the results. However, nouns whose most salient common trait was the lack of sufficient information were consistently grouped together. Thus, the clustering is able to both discriminate between lexical semantic classes and act as a filter to detect those nouns for which there is not sufficient information using only FORMAL role information extracted from corpus data.

## 6 Discussion

As shown, the clustering algorithm discriminated between the three classes considered, using only the FORMAL role descriptors extracted from corpora data as features. Leaving aside the nouns for which there was not enough information available (12.7% of our data set), EVENT, HUMAN and LOCATION nouns were discriminated in the 4-way clustering solution (Clusters 0, 1 and 3 in Table 5, respectively). In this section we analyze misclassified nouns, to understand the reasons behind their misclassification, aiming to evaluate to which extent they correspond to recurring phenomena in language, which can possibly be accounted for by additional strategies.

Although their impact is not significant, noisy extractions (see Section 4.1.1) play a role in misclassification. In the 4-way clustering results, for instance, an EVENT noun is included in the cluster dominated by LOCATION nouns due to errors in extraction, specifically the incorrect identification as a FORMAL role descriptor of the noun in a PP modifying the head noun of an NP, which should be the one extracted. This type of noise is mostly generated by the use of low-level NLP tools. Overall, however, the existence of some noise in the data did not significantly affect the clustering, as demonstrated by the accuracy of the results presented in the previous section.

Concurrently, although general patterns can be identified in language use, one of the main characteristics of language data is its heterogeneity, which means that elements of a given lexical class do not necessarily share all their features or show perfectly matching linguistic behavior. Moreover, considering lexical items are complex objects with different semantic dimensions, they may share properties with elements of more than one lexical class. This type of phenomenon is behind some of the misclassifications in our data, such as the inclusion of *factory*, whose expected lexical class was LOCATION, in the HUMAN nouns cluster. This misclassification seems to be related to the fact that a part of HUMAN class members tended to obtain FORMAL role descriptors typical of HUMAN nouns, as well as of ORGANIZATION nouns, making apparent that nouns do not always occur in the sense considered in our pre-classified list of seed nouns.

## 7 Lexical classes and logical polysemy

As aforementioned, some HUMAN nouns in our list of seed nouns obtain FORMAL role descriptors typical of ORGANIZATION nouns. This is a type of polysemy that occurred in our data only with plural HUMAN nouns, alluding to the work of Copestake (1995) and Caudal (1998), according to whom some HUMAN nouns show a specific type of polysemy when heading definite plural NPs: the polysemy between the individual HUMAN sense and the collection of HUMANs sense, which in turn is polysemous between the HUMANGROUP and ORGANIZATION senses. In (3) we see how the definite plural NP *the doctors* can select for the two senses typically denoted by collective nouns, while having also the possibility to denote individual entities, which is not possible with collectives (see (4a)) that cannot occur in contexts that force a distinct individual entity reading.

    a.   *The <u>doctors</u> lay in the sun.* (several individual HUMAN entities)
    b.   *The <u>doctors</u> protested in front of the hospital.* (HUMANGROUP)
    c.   *The administration negotiated with the <u>doctors</u>.* (ORGANIZATION)    (3)
    a.   # *The <u>staff</u> lay in the sun.* (several individual HUMAN entities)
    b.   *The <u>employees</u> lay in the sun.* (several individual HUMAN entities)
    c.   *The <u>staff</u> protested in front of the hospital.* (HUMANGROUP)
    d.   *The administration negotiated with the <u>staff</u>.* (ORGANIZATION)    (4)

As both collectives and definite plural NPs denote collections, Caudal (1998) states that it is desirable to account for the polysemy of such items morpho-syntactically. This analysis is further strengthened by the observation that, unlike pairs such as *employee* and *staff*, for nouns like *doctor* there is no lexicalization for "group of doctors" in English, the same being true for collective nouns like *audience* or *committee*, whose individual members are not lexicalized. Given such lexical gaps, morpho-syntax is the strategy available. However, though logically polysemous, plural definite NPs like *the doctors* do not allow for multiple selection as is typical of complex types: once the individual HUMAN sense has been selected for there is no access to the HUMANGROUP·ORGANIZATION sense, as suggested by (5) (see Buitelaar (1998) and Rumshisky et al. (2007)).

    *The administration negotiated with the <u>doctors</u>, which later lay in the sun.* (several individual HUMAN entities) (5)

Pustejovsky (1995:155) claims these patterns of linguistic behavior are due to the information in the QS. In the case of expressions like *the doctors,* the dot element denoting the individual HUMAN entity and the complex type HUMANGROUP·ORGANIZATION correspond to different qualia roles, as represented in (6). Hence, the different senses of the expression cannot be selected at the same time.

$$\begin{bmatrix} \text{the doctors} \\ \text{ARGSTR} = \begin{bmatrix} \text{ARG1} = x\text{: human} \\ \text{ARG2} = y\text{: humangroup} \cdot \text{organization} \end{bmatrix} \\ \text{QUALIA} = \begin{bmatrix} \text{FORMAL} = x \\ \text{CONST} = \text{is\_part\_of}(x,y) \end{bmatrix} \end{bmatrix} \quad (6)$$

Going back to the case of *factory*, which was clustered with HUMAN nouns (see Section 6), we will see how the polysemy described above partially applies to this noun. Among the descriptors obtained for *factory* we found, alongside descriptors typical of LOCATION nouns, nouns such as *sector*, *organization* and *profession*, also extracted for HUMAN nouns showing the HUMANGROUP·ORGANIZATION logical polysemy, indicating that nouns like *factory* are also complex objects, as illustrated below by (7):

    a.   *The <u>factory</u> on the corner of Main Street is big and brown.* (LOCATION)
    b.   *The <u>factory</u> summoned a protest against the new government sanctions.* (ORGANIZATION)
    c.   *There was a protest organized* (ORGANIZATION) *by the <u>factory</u> that burned down* (LOCATION) *last week.* (7)

In our data, *factory* shared features both with definite plural NPs headed by HUMAN nouns like *teacher* and *employee* and LOCATION nouns such as *kitchen* and *resort*. The linguistic behavior of *factory* can, therefore, be assumed to reflect the logical polysemy of ORGANIZATION·LOCATION·HUMANGROUP dot types identified by Rumshisky et al. (2007), and represented as follows:

$$\begin{bmatrix} \text{factory} \\ \text{ARGSTR} = \begin{bmatrix} \text{ARG1} = x\text{: location} \\ \text{ARG2} = y\text{: organization} \\ \text{ARG3} = z\text{: human} \end{bmatrix} \\ \text{QUALIA} = \begin{bmatrix} \text{FORMAL} = x \cdot y \cdot z \end{bmatrix} \end{bmatrix} \quad (8)$$

For our work, the most relevant aspect of the behavior displayed by nouns like *factory* is that it makes apparent how our strategy to extract FORMAL role descriptors reflects the ambiguity of nouns to be clustered, which is often difficult to handle in NLP, particularly in classification tasks. The clustering solutions we obtained (see Section 5) grouped together HUMAN nouns, both those that display the ambiguity discussed in this section and those that do not, the same being true for LOCATION nouns. And yet, polysemous nouns display features that clearly point towards the existence of finer-grained distinctions, i.e. sub-classes within lexical semantic classes. This way, particularly given that these fine-grained distinctions are mirrored in FORMAL role descriptors, we assume it should also be possible to automatically recognize groups of nouns within the same ambiguity class, i.e. dot objects.

Hence, we expected the clustering algorithm to identify polysemous lexical items and distinguish them from other members of the same class. To validate this hypothesis we performed an additional iteration of the clustering using the same features and algorithm over previously identified clusters. The iteration was run individually over Clusters 1 and 3 (LOCATION and HUMAN noun clusters, respectively) from our 4-way clustering solution, as both clusters contained logically polysemous nouns. We obtained a 2-way clustering solution for each class, aiming to discriminate nouns strictly containing the LOCATION sense and those reflecting the polysemy described above for *factory*, on one hand, and nouns in the HUMAN·HUMANGROUP·ORGANIZATION ambiguity class from those strictly denoting human individuals on the other. Cluster 1 split into 2 clusters distinguishing between polysemous LOCATION nouns and those that are not, whereas for Cluster 3 the clustering algorithm arrived at a near perfect distinction of dot object nouns and non-ambiguous HUMAN nouns. The noun *factory* clustered with polysemous HUMAN nouns, once more confirming its semantic proximity with nouns of the HUMAN·HUMANGROUP·ORGANIZATION type. Hence, a second iteration of the same clustering algorithm over the same feature vectors was able to identify finer-grained distinctions within lexical classes, automatically recognizing groups of nouns in the same ambiguity class. In doing this, we validate our analysis regarding the role of logical polysemy and dot object types in the clustering solutions obtained, and further strengthen our original hypothesis.

## Final remarks

In this paper, we proposed using automatically obtained FORMAL role descriptors as features to draw together nouns from the same lexical semantic class in an unsupervised clustering task. As there were no available lexica annotated with such information, we obtained it automatically and carried out clustering experiments. In line with the results, our initial hypothesis was supported: in an unsupervised clustering task using FORMAL role descriptors automatically extracted from corpora data as features, we showed it was possible to discriminate between elements of different lexical semantic classes. The filtering and bootstrapping strategy employed proved to minimize effects of sparse data and noise in our task. As shown in the 4-way clustering solution (see Table 5), the clustering exercise, as we designed it, also discriminated the nouns for which there was not sufficient information for a decision to be made on their membership to a cluster corresponding to one of the classes considered. Finally, we explained misclassifications through logical polysemy and showed how the method outlined in this paper allows for making finer-grained distinctions within lexical classes, recognizing lexical items in the same ambiguity class.

The results depicted in this paper demonstrate the validity of our hypothesis, while simultaneously showing that it is possible to incorporate the polysemous behavior of nouns in classification tasks (Hindle, 1990; Bullinaria, 2008) by using an approach that minimizes the effects of sparse data and noise (Bel et al., 2010; 2012). Considering these promising results, in future work we will address the possibility of extending our experiments to other qualia roles, as well as to other lexical semantic classes. At a more applied level, a further step consists in evaluating the feasibility of this approach to automatically extract lexical semantic classes in the automatic acquisition of rich language resources.

## Acknowledgments

This work was funded by the EU 7FP project 248064 PANACEA and the UPF-IULA PhD grant program, with the support of DURSI, and by FCT post-doctoral fellowship SFRH/BPD/79900/2011.